\documentclass[onecolumn,english,linesnumbered, ruled, vlined, boxed]{sig-alternate}
\usepackage[T1]{fontenc}
\usepackage[latin9]{inputenc}
\usepackage{array}
\usepackage{float}
\usepackage{calc}
\usepackage{multirow}
\usepackage{algorithm2e}
\usepackage{amsmath}
\usepackage{amssymb}
\usepackage{graphicx}

\makeatletter

\providecommand{\tabularnewline}{\\}

\usepackage{fixref} % bug fixes

\usepackage{enumitem} % enumeration spacing
\setlist{nolistsep}

\usepackage{balance} % balance last page

\usepackage{tikz}
\usepackage{pgfplots}
\def\@copyrightspace{\relax}

\makeatother

\usepackage{babel}
\begin{document}
\makeatletter
\def\blfootnote{\gdef\@thefnmark{}\@footnotetext}
\makeatother

\title{Attribute Extraction from Product Titles in eCommerce}

\author{\alignauthor{Ajinkya More}\\
\affaddr{@WalmartLabs}\\
\affaddr{860 W California Ave, Sunnyvale CA 94089}\\
\email{amore@walmartlabs.com}}
\maketitle
\begin{abstract}
This paper presents a named entity extraction system for detecting
attributes in product titles of eCommerce retailers like Walmart.
The absence of syntactic structure in such short pieces of text makes
extracting attribute values a challenging problem. We find that combining
sequence labeling algorithms such as Conditional Random Fields and
Structured Perceptron with a curated normalization scheme produces
an effective system for the task of extracting product attribute values
from titles. To keep the discussion concrete, we will illustrate the
mechanics of the system from the point of view of a particular attribute
- brand. We also discuss the importance of an attribute extraction
system in the context of retail websites with large product catalogs,
compare our approach to other potential approaches to this problem
and end the paper with a discussion of the performance of our system
for extracting attributes. 
\blfootnote{
Copyright is held by the author/owner(s)\\
Workshop on Enterprise Intelligence, Aug 14, KDD 2016
 }
\end{abstract}

\section{Introduction}

\subsection{Vocabulary}

Before beginning the discussion of the problem we will first define
some terms that will be used in the remainder of the paper. A \emph{product
}is any commodity which may be sold by a retailer. An \emph{attribute
}is a feature that describes a specific property of a product or a
product listing. Some examples of attributes include brand, color,
gender, material, title, description, etc. An \emph{attribute value
}is a particular value assumed by the attribute. For example, for
the product title 
\begin{quote}
\textsf{\scriptsize{}Apple iPad Mini 3 16GB Wi-Fi Refurbished, Gold}{\scriptsize \par}
\end{quote}
the brand attribute value is \emph{`Apple' and }the color attribute
value is \emph{`Gold'. A} product may alternatively be defined as
a collection of such attribute-value pairs, where a value can potentially
be empty. Formally, let $\mathfrak{p}$ be a product with attributes
$\alpha_{1},\alpha_{2},...,\alpha_{m}$ and values $v_{1},v_{2},...,v_{m}$
respectively. Then, we represent $\mathfrak{p}=\{\alpha_{1}:v_{1},\alpha_{2}:v_{2},...,\alpha_{m}:v_{m}\}$
and write $\mathfrak{p}(\alpha_{i})=v_{i}$ for $1\le i\le m$ to
indicate an attribute-value relationship for $\mathfrak{p}$. 

In the context of an eCommerce website, the attribute values may be
used to filter search results based on the items with a matching attribute
value. As is done on Walmart.com and several other retail websites,
this may be accomplished by populating relevant attribute values in
the left hand navigation pane. For any given search or browse session,
the attributes that are displayed for further filtering in the left
hand navigation as described above will be called \emph{facets.}

For the sake of brevity, we use the terms `attribute value extraction'
and `attribute extraction' synonymously.

\subsection{Problem set up\label{sub:Problem-set-up}}

We formalize attribute extraction as per the following definition.

\newdef{definition}{Definition} 

\begin{definition}Let $x$ be a product title and let $(x_{1},x_{2},..,x_{n})$
be a particular tokenization $\mathbf{x}_{\mathbf{t}}$ of $x$. Given
an attribute $\alpha$, \emph{attribute extraction }is the process
of discovering two functions $E$ (raw extraction) and $N$ (normalization)
such that 
\begin{itemize}
\item $E(\mathbf{x_{t}})=E((x_{1},x_{2},...,x_{n}))=(x_{i},x_{i+1},...,x_{k})$
for $1\leq i\leq k\leq n$ where $\mathbf{a}_{\mathbf{v}}=(x_{i},x_{i+1},...,x_{k})$
is a tokenization of a particular value of $\alpha$,
\item $N((x_{i},x_{i+1},...,x_{k}))=a_{s}$ where $a_{s}$ is the standardized
representation of \textbf{$\mathbf{a_{v}}$}.
\end{itemize}
\end{definition}

\newdef{example}{Example} 

\begin{example}

Consider the product title
\begin{quote}
$x=$\textsf{\scriptsize{} Hewlett Packard B4L03A\#B1H Officejet Pro
Eaio}{\scriptsize \par}
\end{quote}
so that whitespace tokenization yields 
\[
\mathbf{x_{t}}=(x_{1},x_{2},...,x_{6})=(\text{Hewlett, Packard, ..., Eaio}).
\]
 Let $\alpha$ be the attribute \textbf{`}brand'. Then we seek to
find two functions $E$ and $N$ such that 
\[
E((x_{1},x_{2},...,x_{6}))=(x_{1},x_{2})=(\text{Hewlett, Packard})
\]
and 
\[
N((\text{Hewlett, Packard}))=HP.
\]

\end{example}

The goal of the attribute extraction system is to minimize the \emph{loss
function} $1-F_{1}$ where the $F_{1}$ measure is as defined in section
\ref{sub:Metrics}.

Attribute value extraction is a particular instance of a named entity
recognition problem. This paper explores the use of machine learning
techniques, in particular, sequence labeling algorithms for the purposes
of extracting attribute values from product titles. We will also illustrate
a normalization scheme which provides the dual benefits of standardizing
variations in the same attribute value and boosting precision. 

In section \ref{SEC: CHALLENGES}, we outline some of the challenges
associated with attribute extraction in general. We reference some
of the previous work on the problem of entity extraction in section
\ref{SEC: PRIOR-WORK}. Section \ref{SEC: SLA} describes the sequence
labeling algorithms used to build an attribute extraction system that
works well with eCommerce data. This system is described in detail
in sections \ref{SEC:ANNOTATING} - \ref{SEC:RESULTS} with respect
to a particular attribute - `brand'. We finish the paper by describing
how similar approaches have been successful in extracting other attributes
such as 'character', `manufacturer part number', `package quantity',
etc and discussing the importance of attribute extraction in eCommerce.

\section{Use Cases \label{SEC:IMPACT}}

In this section, we elaborate on the importance of attribute extraction
for retail websites.

\subsection{Discoverability}

Using facets to filter search or browse results is a common way for
consumers of eCommerce websites to navigate the site and search through
the product space. In order to ensure a good faceted navigation experience,
it is critical to associate attribute value metadata to products for
the attributes that appear in the facets. 

For example, let $S$ be a search query entered by a user and $\mathcal{R}$
be the set of products returned as a result. Suppose product $\mathfrak{p}\in\mathcal{R}$
and $\mathfrak{p}$ has an attribute $\alpha$ which also happens
to be a facet. Suppose further that the value of $\alpha$ applicable
to $\mathfrak{p}$ is $v$. When the user clicks on facet value $v$,
the filtered result set is $\mathcal{R}^{'}\subseteq\mathcal{R}$.
Then $\mathfrak{p}\in\mathcal{R}^{'}$ if and only if $\mathfrak{p}(\alpha)=v$.
More concretely, suppose a user enters the search query `Tee shirt'
and the result set contains the product titled
\begin{quote}
\textsf{\scriptsize{}Hanes Mens NANO-T Dri T-Shirt S Deep Red}{\scriptsize \par}
\end{quote}
However, if the attribute `color' for the product is missing the value
`Red', then if the user clicks on the value `Red' under facet `color',
this product will no longer show in the result set even though it
has the attribute value of interest to the user.

Thus, absence of relevant attribute values in product metadata has
a direct bearing on the discoverability of the product and ultimately
funnels down to affect the sale of such products. 

It is crucial for modern eCommerce sites to be able to add products
to its catalog as quickly as possible. As such, in order to ensure
that the item set up process has minimal requirements, providing attribute
values may not be enforced for all attributes. Consequently, a large
fraction of incoming product data may not have attribute values supplied.
In the absence of a system to tag missing attribute values, the discoverability
of such products will be diminished.

\subsection{Ad campaigns}

Certain attribute values are also are required by ad publishers such
as Google and Bing in order to launch ad campaigns for products. In
the absence of the required attributes, the ad campaigns are rejected.

\subsection{Compliance}

Some attribute values are necessary to satisfy government compliance
requirements, e.g. unit price for food items or items sold in bulk.

\subsection{Knowledge discovery}

The list of valid values for a particular attribute may not be fixed.
This is true, for instance, with attributes like `brand', `character',
`model number', etc. It is desirable to build a solution that can
discover attribute values not currently part of the knowledge base.

\section{\label{SEC: CHALLENGES}Challenges}

\subsection{Extracting attributes from product titles}

We will describe a sequence labeling based machine learning system
for extracting values of certain product attributes from product titles.
We will discuss the challenges associated with this task and how they
were handled in the current solution. In the next few sections we
will describe the attribute extraction system using the specific example
of brand extraction. 

In certain product categories, brand is an important attribute. Extracting
brands from product titles presents several interesting challenges,
some of which are outlined below. Whenever possible, we include examples
of actual titles of products sold on Walmart.com. In later sections
we discuss how our solution mitigates these issues.
\begin{itemize}
\item Unlike English prose, product titles do not adhere to a syntactic
structure. They may be a concatenation of several nouns and adjectives
as well as product specific identifiers and acronyms. Verbs tend to
be missing and there is no standardized way of handling letter case.
For example, consider the following titles of actual Walmart products
(the brand names are in bold).

\begin{itemize}
\item \textsf{\scriptsize{}Chihuahua Bella Decorative Pillow by }\textsf{\textbf{\scriptsize{}Manual
Woodworkers and Weavers}}\textsf{\scriptsize{} - SLCBCH}{\scriptsize \par}
\item \textsf{\textbf{\scriptsize{}Real Deal Memorabilia}}\textsf{\scriptsize{}
BCosbyAlbumMF Bill Cos}{\scriptsize \par}
\end{itemize}
\item Due to the diversity of products sold in any leading eCommerce site,
product titles do not follow any specific composition. For instance,
the location of brands within titles may vary. Additionally, the number
of tokens that constitute a brand name is also highly variable.

\begin{itemize}
\item \textsf{\scriptsize{}Old World Prints OWP86575Z Romantic Jasmine Poster
Print by }\textsf{\textbf{\scriptsize{}Vision studio}}\textsf{\scriptsize{}
- 18 x 22}{\scriptsize \par}
\item \textsf{\textbf{\scriptsize{}Autograph Warehouse}}\textsf{\scriptsize{}
84377 Jake Rodriguez Card Boxing 1996 Ringside No . 40}{\scriptsize \par}
\item \textsf{\scriptsize{}Straight Talk }\textsf{\textbf{\scriptsize{}Samsung}}\textsf{\scriptsize{}
Galaxy S3 Prepaid Cell Phone, White}{\scriptsize \par}
\end{itemize}
\item Further, different products may contain slightly varying spellings
of the same brand. This may include presence or absence of spaces
and hyphens, presence or absence of apostrophe, presence or absence
of trailing words such as `inc.' or `ltd.', etc.

\begin{itemize}
\item \textsf{\textbf{\scriptsize{}J\&C Baseball Clubhouse}}\textsf{\scriptsize{}
JC000213 WWE John Cena Engraved Collector Plaque with 8x10 KNOCK OUT
Photo}{\scriptsize \par}
\item \textsf{\textbf{\scriptsize{}J \& C Baseball Clubhouse}}\textsf{\scriptsize{}
JC000008 Pittsburgh Penguins All Time Greats 6 Card Collector Plaque}{\scriptsize \par}
\end{itemize}
\item Some titles may contain abbreviations of brand names.

\begin{itemize}
\item \textsf{\textbf{\scriptsize{}Kcl}}\textsf{\scriptsize{} 2SL25WH Accessory
LED Tape Power Supply Lead in White}{\scriptsize \par}
\item \textsf{\textbf{\scriptsize{}Kichler}}\textsf{\scriptsize{} Builder
5019NI 8 Light Bath Strip in Brushed Nickel}{\scriptsize \par}
\end{itemize}
\item Brand names in titles may contain typographical errors.

\begin{itemize}
\item \textsf{\textbf{\scriptsize{}Trademak Global}}\textsf{\scriptsize{}
AD-CLC4000-PITT Pittsburgh Panthers 40 inch Rectangular Stained Glass
Billiard Light}{\scriptsize \par}
\item \textsf{\textbf{\scriptsize{}Trademark Global}}\textsf{\scriptsize{}
24\textquotedbl{} Cushioned Folding Stool}{\scriptsize \par}
\end{itemize}
\item A case of particular interest is that of generic or unbranded products.
There is especially a preponderance of such products that fall under
jewelry and clothing categories.

\begin{itemize}
\item \textsf{\scriptsize{}0.5Ctw Diamond Fashion Womens Fixed Ring Size
- 7}{\scriptsize \par}
\item \textsf{\scriptsize{}Women's Popcorn Stitch Infinity Scarf}{\scriptsize \par}
\end{itemize}
\item There are categories of products for which brand name is not an important
attribute. In such cases, the brand facet will not even be displayed
for a search of items belonging to these categories. Examples of such
categories include books, movie DVDs, posters, etc.
\item The list of brand names relevant to a given product catalog is constantly
changing. Products with new brand names may appear, while some old
brands may no longer sell any products. Thus, it is desirable for
the brand extraction algorithm to be able to discover new brand names
as well as extract known brands with high precision.
\item Collecting expert feedback either for the purposes of generating training
data or validating model generated labels is subject to inter-annotator
disagreement. It is not ideal to show different facet values that
correspond to the same brand as it diminishes the quality of user
experience. This is one of the reasons that makes crowd sourcing an
unsatisfactory solution to the problem of attribute extraction.
\end{itemize}

\subsection{Comparison with other approaches}

\subsubsection{Dictionary based lookup }

A simple method for attribute extraction is to prepare a curated lexicon
of attribute values and given a product title, scan it to find a value
from the list. Some gazetteer based approaches are discussed in \cite{nadeau2007survey}
in the context of named entity recognition problems. This approach
suffers from numerous drawbacks. The curated list will need to be
constantly updated in order to match to new attribute values in products.
For certain attributes, the number of possible values can be of the
order of the number of products themselves, \emph{e.g. `}manufacturer
part number'. For such attributes, lookup based approaches are completely
ineffective. Further, as mentioned earlier, the same attribute value
may occur in a variety of different forms in the title, so the curated
list will need to discover and keep track of all variations. Finally,
such a system will need to devise a mechanism to break ties in case
of multiple matches.

\subsubsection{Crowd Sourcing}

Crowd sourcing as a solution for extracting attributes for products
is rendered ineffective because of the scale of a major retail catalog.
Given the large number of attributes that need to be extracted, potentially
for millions of products, the time and cost of crowd sourcing make
it prohibitive. In addition, since variations of attribute values
may need to be standardized, resolving inter annotator disagreement
can be challenging and may require expert intervention.

\subsubsection{Rule based extraction}

Rule based approaches have had success in certain named entity recognition
tasks \cite{rule-other}, \cite{rule-greek}, \cite{mikheev1999named}.
Such techniques typically leverage the grammatical structure of the
language. However, as mentioned earlier in the section, product titles
do not conform to a syntactical structure or grammar unlike news articles
or prose. An alternative would be to use rule based approaches with
product description, but descriptions may contain named entities unrelated
to the product such as in comparisons to similar products. 

Another disadvantage with rule based approaches stems from the number
of different attributes that are used by a general retailer to describe
products. Typically, a modern retailer deals with tens of thousands
of attributes across thousands of product categories. Rule based approaches
will need to be tailored for each attribute. Creating and maintaining
rules for hundreds or thousands of attributes can be quite challenging.
In contrast we were able to easily adapt our system with minimal changes
to build models for a variety of attributes.

\subsubsection{Supervised text classification}

Yet another possibility is to use text classification algorithms such
as logistic regression, naive bayes or support vector machines that
do not leverage the sequential structure of the data. These algorithms
can be suitable for certain attributes, where the number of classes
is known and small. In this scenario, classification algorithms can
provide great performance. In contrast, when the number of classes
is in tens of thousands, we will need a lot labeled training data
and the model footprint will also be large. However, the main drawback
with these models for attributes like brand and manufacturer part
number is that they can only predict classes on which they are trained.
Thus, in order to predict new brand values, the training data will
need to be constantly updated with labeled data corresponding to new
brands. In the case of manufacturer part number, this approach is
essentially worthless since every new product will likely have an
unseen part number.

\section{Prior Work \label{SEC: PRIOR-WORK}}

Tagging products with attributes falls under the umbrella of knowledge
extraction from text. Such problems have been explored in a variety
of application areas. We will present a brief survey of similar undertakings
in other domains. 

Bikel \emph{et.al. }\cite{NER} discuss the named entity extraction
problem to identify location names, person names, named organizations
and a few other entities. They present a Hidden Markov model and show
that it performs favorably on named entity recognition tasks on standard
datasets like MUC-6 and MET-1. Ritter \emph{et. al. }\cite{tweets}
built a system for mining named entities from tweets - which is another
example of text that significantly deviates from fluent prose. 

Part of speech tagging is a quintessential example of an entity recognition
task and a number of approaches have been investigated in literature.
Eric Brill \cite{postag} introduced a simple rule based tagger for
learning part of speech which was shown to have an error rate of 7.9\%
when trained on 90\% of Brown corpus and tested on a held out 5\%
subset. Schmid \cite{schmid} illustrates a probabilistic tagger in
which transition probabilities are estimated using a decision tree
and which achieves an accuracy of 96.36\% on Penn-Treebank data. Ratnaparkhi
describes a maximum entropy approach to this problem that achieves
an accuracy of 96.6\% on Wall Street Journal data in Penn-Treebank.

Alani \emph{et. al. }\cite{ontology} present a system to extract
knowledge about artists from web pages leveraging a curated ontology.
Kazama and Torisawa \cite{wiki} use features relying on Wikipedia
information to generate a Conditional Random Fields (CRF) based model
for named entity recognition. Etzioni \emph{et. al. }\cite{unsupervised}
describe unsupervised techniques to extract relationships between
entities from web documents. 

The problems that are the subject of this paper are most similar to
the following works. Ghani \emph{et. al. }\cite{ghani}\emph{ }discuss
the representation of retail products as attribute-value pairs. They
tackle the problem of extracting values for a predefined list of attributes
like age group, degree of brand appeal and price point. After obtaining
an initial expert labeled dataset, they augment it with unlabeled
data using Expectation-Maximization. Popescu and Etzioni \cite{reviews}
extend their system described in \cite{unsupervised} to the problem
of mining product features from online reviews. Putthividhya and Hu
\cite{ebay} designed a system for extracting product attributes from
short listing titles such as those found on eBay. They focus on specific
categories - clothing and shoes and on the attributes brand, style,
size and color. They compare performance of Hidden Markov models,
Maximum Entropy models, Support Vector Machines and CRFs. Beginning
with a seed data set of labeled attribute values, they generate additional
training data using unlabeled examples. For normalizing variations
in the extracted attribute values, they use n-gram substring matching.

\section{Sequence Labeling Approaches\label{SEC: SLA}}

We propose a sequence labeling based approach for identifying brands
from product titles. Given an input sequence $\mathbf{x}=x_{1},x_{2},...,x_{m}$,
a sequence labeling algorithm aims to unearth a label sequence $\mathbf{y}=y_{1},y_{2},...,y_{m}$
such that the element $x_{j}$ is labeled $y_{j}$. The labels $y_{j}$
usually come from a finite set. A typical example of such an approach
is part of speech tagging in natural language processing. 

In this work we evaluated performance using two sequence labeling
algorithms - Structured Perceptron and Conditional Random Fields.
In section \ref{SEC:RESULTS} we compare the performance of these
models with respect to another sequence labeling algorithm - Hidden
Markov model.

\subsection{Feature Functions \label{SUB: FEATURE-FUNCTIONS}}

Let $X$ be the set of all input sequences and let $Y$ be the set
of all label sequences of length $m$. Let $I=\{1,2,...,m\}$. A feature
function for a sequence labeling algorithm is a function $f:X\times Y\times I\to\mathbb{R}$.

Consider for example the problem of part of speech tagging. This involves
assigning every word in a unit of text (\emph{e.g. }a sentence) a
tag/label corresponding to its part of speech. The tags that appear
in Penn Treebank \cite{penntb} include DT (determiner), JJ (adjective),
NN (noun), VB (verb) and IN (preposition). Now let $\mathbf{x}=$
(The, quick, brown, fox, jumps, over, the, lazy, dog), $\mathbf{y}=$
(DT, JJ, JJ, NN, VB, IN, DT, JJ, NN). We may define a feature function
as follows:
\begin{align*}
f(\mathbf{x},\mathbf{y},i)=\begin{cases}
1 & \text{if }x_{i}=\text{the and }y_{i}=DT\\
0 & \text{otherwise}
\end{cases}.
\end{align*}
Then $f(\mathbf{x},\mathbf{y},2)=0$ and$f(\mathbf{x},\mathbf{y},7)=1$. 

We may also define feature functions that take into account contextual
information of tokens in the input sequence. In fact, such features
are critical to the success of a sequence labeling algorithm, for
otherwise, we might as well build a traditional classifier ignoring
the sequence structure. An example of such a feature function is given
below:

\begin{align*}
f(\mathbf{x},\mathbf{y},i)=\begin{cases}
1 & \text{if }x_{i}\text{ is capitalized and }x_{i+1}\text{ is }\\
 & \text{not capitalized}\\
0 & \text{otherwise}
\end{cases}.
\end{align*}
Note that $f(\mathbf{x},\mathbf{y},1)=1$ and $f(\mathbf{x},\mathbf{y},i)=0$
for $i\neq1$.

Suppose we construct $d$ feature functions $f_{1},f_{2},...,f_{d}$.
Let 
\[
F_{i}(\mathbf{x},\mathbf{y})=(f_{1}(\mathbf{x},\mathbf{y},i),f_{2}(\mathbf{x},\mathbf{y},i),...,f_{d}(\mathbf{x},\mathbf{y},i))
\]
be the $d$-dimensional feature vector corresponding to the to the
pair $\mathbf{x},\mathbf{y}$ and position $i$. We denote 
\[
F(\mathbf{x},\mathbf{y})=\Sigma_{i=1}^{m}F_{i}(\mathbf{x},\mathbf{y})
\]
 as the $d$-dimensional feature vector corresponding to the pair
$\mathbf{x},\mathbf{y}$. Note that although $F$ depends on the length
of the sequence $m$, for a given input sequence $\mathbf{x}$, we
evaluate $F$ over pairs $(\mathbf{x},\mathbf{y})$ where $\mathbf{y}$
varies over candidate label sequences which will all have the same
length as the length of $\mathbf{x}$.

We will discuss the feature functions used in building models for
the brand extraction algorithms in section \ref{SEC:FEATURES}.

\subsection{Structured Perceptron}

Michael Collins presented the \emph{Structured Perceptron} learning
algorithm in \cite{collins}. We use the refinement of the algorithm
called \emph{averaged parameters} in that paper. Parameter averaging
reduces variance providing a regularization effect and improves performance
(\cite{collins}, \cite{avg-perc}).

Structured Perceptron is a supervised learning algorithm. The training
set to the algorithm consists of labeled sequences $\{(\mbox{\textbf{x}}_{i},\mbox{\textbf{y}}_{i})\}$
where $i=1,2,...,n$. Each input $\mathbf{x}_{i}$ is a sequence of
the form $(x_{1},x_{2},...,x_{m})_{i}$ with a corresponding sequence
of labels $(y_{1},y_{2},...,y_{m})_{i}$ such that the input sequence
element $x_{j}$ has a corresponding label $y_{j}$. The labels belong
to a finite set $Y_{L}$. Let $Y_{S}$ denote the set of all sequences
of length $m$ such that each entry in the sequence belongs to $Y_{L}$.
Thus, $|Y_{S}|=|Y_{L}|^{m}$. 

Let $d$ is the number of feature functions as illustrated in section
\ref{SUB: FEATURE-FUNCTIONS}. We outline the Structured Perceptron
with averaged parameters (SP) algorithm below.

\medskip{}

\fbox{\begin{minipage}[t]{0.9\columnwidth}%
\begin{enumerate}
\item Initialize $\mbox{\textbf{w}}=(0,0,...0)$ and $\mbox{\textbf{w}}_{a}=(0,0,...0)$
where $\mbox{\textbf{w}},\mbox{\textbf{w}}_{a}$ are tuples of length
$d$
\item for $j=1$ to $N$
\item \qquad{}for $i=1$ to n
\item \qquad{}\qquad{}$\mathbf{y}_{i}^{*}=$ $\text{arg max}_{\mathbf{y}\in Y_{S}}\mathbf{w}^{T}F(\mathbf{x}_{i},\mathbf{y}_{i})$
\item \qquad{}\qquad{}if $\mathbf{y}_{i}\neq\mathbf{y}_{i}^{*}$, then
\item \qquad{}\qquad{}\qquad{}$\mathbf{w}=\mathbf{w}+F(\mathbf{x}_{i},\mathbf{y}_{i})-F(\mathbf{x}_{i},\mathbf{y}_{i}^{*})$
\item \qquad{}\qquad{}$\mathbf{w}_{a}=\mathbf{w}_{a}+\mathbf{w}$
\item Return $\frac{\mathbf{w}_{a}}{nN}$\end{enumerate}
\end{minipage}}

\medskip{}

Step 4 of the algorithm is implemented using Viterbi decoding.

\subsection{Linear Chain Conditional Random Fields}

Conditional Random Fields (CRF) is a probabilistic structured labeling
algorithm introduced by Lafferty, McCallum and Pereira in \cite{crf}.
In our system, we used a linear chain CRF (LCCRF). As before, let
$\mathbf{x},\mathbf{y}$ denote input and label sequences and let
$Y_{S}$ represent the set of all label sequences. Let $F(\mathbf{x},\mathbf{y})$
be the $d$-dimensional feature vector corresponding to the pair $\mathbf{x},\mathbf{y}$
as defined in section \ref{SUB: FEATURE-FUNCTIONS}. Then, for a given
weight vector $\mathbf{w}\in\mathbb{R}$, according to the LCCRF model
we have, 
\[
\text{Pr}(\mbox{\textbf{y}}=\mbox{\textbf{y}}^{*}|\mathbf{x};\mathbf{w})=\frac{\text{exp}(\mathbf{w}^{T}F(\mathbf{x},\mathbf{y}^{*}))}{\Sigma_{\mathbf{y}\in Y_{S}}\text{exp}(\mathbf{w}^{T}F(\mathbf{x},\mathbf{y}))}.
\]

Given a weight vector, we seek to find a label sequence that maximizes
the above conditional probability. We may simplify this as follows
\begin{align*}
 & \text{arg max}_{\mathbf{y}\in Y_{S}}\text{Pr}(\mbox{\textbf{y}}|\mathbf{x};\mathbf{w})\\
 & =\text{\text{arg max}}_{y\in Y_{S}}\frac{\text{exp}(\mathbf{w}^{T}F(\mathbf{x},\mathbf{y}^{*}))}{\Sigma_{\mathbf{y}\in Y_{S}}\text{exp}(\mathbf{w}^{T}F(\mathbf{x},\mathbf{y}))}\\
 & =\text{arg max}_{y\in Y_{S}}\text{exp}(\mathbf{w}^{T}F(\mathbf{x},\mathbf{y}^{*}))\\
 & =\text{arg max}_{y\in Y_{S}}\mathbf{w}^{T}F(\mathbf{x},\mathbf{y}^{*})
\end{align*}
As before we search for the sequence maximizing the above dot product
using Viterbi decoding. 

Finally we need to estimate the weight vector $\mathbf{w}$. Suppose
we have a set of labeled sequences $\{(\mbox{\textbf{x}}_{i},\mbox{\textbf{y}}_{i})\}$
where $i=1,2,...,n$ as our training set. We define the conditional
log-likelihood of the data as
\[
L(\mathbf{w})=\Sigma_{i=1}^{n}\text{log}\text{Pr}(\mbox{\textbf{y}}_{i}|\mathbf{x}_{i};\mathbf{w})
\]
and the $L2-$regularized log likelihood as 
\[
L_{2}(\mathbf{w})=\Sigma_{i=1}^{n}\text{log}\text{Pr}(\mbox{\textbf{y}}_{i}|\mathbf{x}_{i};\mathbf{w})-\frac{\lambda}{2}||\mathbf{w}||^{2}.
\]

The parameter estimation problem is now posed as an optimization problem
as follows
\[
\mathbf{w}^{*}=\text{arg max}_{\mathbf{w}\in\mathbb{R}^{d}}L_{2}(\mathbf{w}).
\]

\section{Annotating Product Titles\label{SEC:ANNOTATING}}

\subsection{\label{SUB:LABELING}Labeling scheme}

As an input for both models, we used a product title decomposed into
a sequence of tokens labeled with BIO encoding. The first step of
the labeling process tokenizes the title into a sequence of tokens
separated by white space and/or certain special characters. The BIO
encoding scheme assigns one of the following three labels to each
of the tokens obtained this way.
\begin{enumerate}
\item \textbf{B-brand}: This label indicates that the token is the beginning
token of a brand name. It will not be used if the title does not contain
a brand name.
\item \textbf{I-brand}: This label indicates that the token is an intermediate
(\emph{i.e. }any token other than the first) token of a brand name.
It will not be used if the title does not contain a brand name or
if the brand name in the title has a length of 1 token.
\item \textbf{O}: Indicates that the token is not part of a brand name in
the corresponding title.\textbf{ }
\end{enumerate}
We illustrate this encoding on a product title with the brand name
\emph{`}Manual Woodworkers and Weavers'.

\medskip{}

$\begin{aligned}\underbrace{\text{Manual}}_{\text{B-brand}}\ \underbrace{\text{Woodworkers}}_{\text{I-brand}}\ \underbrace{\text{and}}_{\text{\text{I-brand}}}\ \underbrace{\text{Weavers}}_{\text{\text{I-brand}}}\end{aligned}
$

$\begin{aligned}\underbrace{\text{AIMFBQ}}_{\text{O}}\ \underbrace{\text{Butterfly}}_{\text{O}}\ \underbrace{\text{Quilt}}_{\text{O}}\ \underbrace{\text{32X42}}_{\text{O}}\ \underbrace{\text{inch}}_{\text{O}}\end{aligned}
$

\subsection{Distantly supervised training data}

In order to capture the variations in which an attribute name may
appear in product titles, we needed a sizable training set. However,
generating a reasonable sized training data set using manual labeling
would require substantial tedious and time consuming effort. Alternatively,
we could use the set of products which were already tagged with the
corresponding attribute value. However, this approach is susceptible
to any noise in the existing tags which was found to be significant. 

We used a distant supervision approach to build our initial training
data set. For each attribute we built regex based rules to programatically
annotate product titles as per the labeling scheme in section \ref{SUB:LABELING}.
For example, in the case of brand, we only annotated product titles
in which the brand name appears exactly as a substring, except possibly
for a change in the case and/or the presence or absence of certain
special characters. Note that not all brand values appear exactly
in product titles owing to value normalization which we discuss in
more detail in section \ref{SUB:NORMALIZATION}. For example the following
title 
\begin{quote}
\textsf{\textbf{\scriptsize{}Kimberly-Clark}}\textsf{\scriptsize{}
Nitrile Xtra Exam Medium Gloves in Purple}{\scriptsize \par}
\end{quote}
with an associated brand attribute value of `\emph{Kimberly-Clark'
}could be included in the training set. However, the title 
\begin{quote}
\textsf{\scriptsize{}Barnett Single Friction Plate Fits 76-79 Yamaha
RD400}{\scriptsize \par}
\end{quote}
where the normalized Walmart brand value is `\emph{Barnett Crossbows}'
would not be included since the word `Crossbows' does not appear in
the title. In particular, the initial training set did not contain
any product titles where the product had a legitimate brand value
but which was not contained in the title. 

This restriction achieved the dual purpose of shielding the training
set from noisy labels as well as allowing us to programmatically annotate
product titles without the need for any manual labeling. For each
such brand name, we included up to three annotated titles to the training
set. The reason for this limit was to avoid having a heavily imbalanced
dataset since the distribution of the number of items with a given
brand name was observed to be heavy tailed.

Note that although the initial training set did not contain any product
titles for which the normalized brand value did not appear in the
product title, we were subsequently able to increase the size of the
training set with such product titles from the analyst feedback we
received during the validation phase (see section \ref{SUB:MANUAL}).

In addition, we also wanted the algorithm to detect product titles
which do not contain any brand name. A couple of examples of such
titles were provided in Section \ref{SEC: CHALLENGES}. To this end,
we added a small set of such product titles corresponding to unbranded
products where every token was labeled `O'.

\subsection{Interpreting output labels}

The output of the learning algorithm on a product title $x$ is a
sequence of labels - one label per token in the tokenization $\mathbf{x_{t}}$
of $x$, according to the BIO-encoding scheme specified in section
\ref{SUB:LABELING}. This labeling is transformed into a candidate
brand name using the following steps:
\begin{enumerate}
\item If the label of all tokens of the product title is `O', we posit that
a brand name does not appear in the title and output value for the
product title is \emph{`unbranded'}. In the notation of section \ref{sub:Problem-set-up},
\[
E(\mathbf{x_{t})}=\text{`'}\text{ and }N(E(\mathbf{x_{t}))=}\text{ `unbranded'}.
\]

\item Otherwise, there must be a token with the label `\emph{B-brand'}.
Consider the contiguous subsequence of tokens $\mathbf{x_{s}}$ satisfying:

\begin{enumerate}
\item The label of the first token is `\emph{B-brand'. }
\item If the length of the subsequence is greater than 1, then the label
of each token except the first is `\emph{I-brand'.}
\item The last token in the subsequence is either the last token of the
product title or the succeeding token from the product title is labeled
`\emph{O'.}
\end{enumerate}
\end{enumerate}
The substring of the product title corresponding to $\mathbf{x_{s}}$
(if applicable) is considered to be the predicted brand for $x$ \emph{i.e.
$E(\mathbf{x_{t}})=\mathbf{x_{s}}$.} We discuss normalization for
this case in section \ref{SEC:POSTPROCESSING}.

Note that each product title corresponds to a unique brand. Thus,
every label sequence in the training data consists of at most one
`\emph{B-brand' }label. In practice, label sequences produced by the
algorithm also consist of a single `\emph{B-brand' }label. In the
rare case that multiple such sequences exist in a given product title,
we deem the predicted brand value to be the one corresponding to the
first such subsequence. This heuristic works well with the datasets
we analyzed. Depending on the attribute a different strategy for resolving
ties may be necessary. Alternately, in the case of multi valued attributes,
we may accept all possible subsequences satisfying the above criteria.

\section{Features\label{SEC:FEATURES}}

Both SP and LCCRF models were trained using a similar set of features.
The choice of feature functions was motivated from a variety of literature
on this subject including \cite{features} and \cite{ebay}. We curated
and adapted the feature functions to those that improved the performance
of the attribute extraction system. Feature selection was done using
ablative analysis. In this methodology, we start with a set of feature
functions and measure the impact of iteratively turning off each feature
on relevant metrics (e.g. $F_{1}$ measure). The final set of feature
functions used has the property that removing any feature function
adversely impacts the metric of interest. Alternately, we can start
with one feature function and iteratively keep adding feature functions
that improve performance.

\subsection{Feature functions}

Some examples of features used in the attribute extraction system
are given below. In particular, unlike many other named entity recognition
systems, we don't employ any features based on attribute specific
lexicons.

\subsubsection{Characteristic features}

These features are derived from information about a given token alone.
For example the identity of the token, the character composition of
the token, letter case, token length in terms of number of characters.

\subsubsection{Locational features}

These features depend on the position of the token in the token sequence
into which the title is decomposed. For instance, number of tokens
in the title before the given token.

\subsubsection{Contextual features}

In order for sequence labeling algorithms to work well, we need to
capture information about tokens neighboring a given token. This may
be achieved using features such as the identity of the preceding/succeeding
token, whether the preceding/succeeding token is capitalized, the
bigram consisting of the token and its predecessor/successor, whether
the preceding token is a conjunction, part of speech tag of the token,
etc.

\subsection{Features used in brand extraction}

For the particular case of brand extraction, we show below the set
of features that gave the best performance over a sample of product
titles in selected departments. We progressively added features until
there was an improvement in the average $F_{1}$ measure%
\footnote{We formally define the relevant metrics in the context of attribute
extraction in section \ref{sub:Metrics}.%
} obtained from a 10-fold cross validation. 

We make use of the following notation. To determine feature functions
corresponding to a position $i$ in a title, let $w_{0}$ denote the
token under consideration. If applicable, let $w_{-1},w_{-2},...$
denote the preceding tokens and let $w_{1},w_{2},...$ denote the
successive tokens. For a token $w$ let $w[j]$ denote the $j$th
character of $w$. 

The table below shows the drop in average $F_{1}$ measure, $\Delta F_{1}$
(in percentage), recorded by turning off a given feature at a time.
All measurements are based on a 10-fold cross validation. A `-' against
a feature function indicates it was not used for that model.

\medskip{}

\textsf{\scriptsize{}}%
\begin{tabular}{|c|c|c|}
\hline 
\textsf{\textbf{\scriptsize{}Feature function}} & \multirow{1}{*}{\textsf{\textbf{\scriptsize{}$\Delta F_{1}$ (LCCRF)}}} & \textsf{\textbf{\scriptsize{}$\Delta F_{1}$ (SP)}}\tabularnewline
\hline 
\hline 
\textsf{\scriptsize{}$(w_{-1},w_{0})$} & \textsf{\scriptsize{}0.577} & \textsf{\scriptsize{}0.675}\tabularnewline
\hline 
\textsf{\scriptsize{}$(w_{0},w_{1})$} & \textsf{\scriptsize{}0.423} & \textsf{\scriptsize{}0.443}\tabularnewline
\hline 
\textsf{\scriptsize{}$w_{0}$} & \textsf{\scriptsize{}0.220} & \textsf{\scriptsize{}0.181}\tabularnewline
\hline 
\textsf{\scriptsize{}$w_{-1}$ lemma} & \textsf{\scriptsize{}0.177} & \textsf{\scriptsize{}0.079}\tabularnewline
\hline 
\textsf{\scriptsize{}$(w_{-2},w_{-1})$} & \textsf{\scriptsize{}0.207} & \textsf{\scriptsize{}0.272}\tabularnewline
\hline 
\textsf{\scriptsize{}$w_{1}[0]$ is a digit} & \textsf{\scriptsize{}0.146} & \textsf{\scriptsize{}0.120}\tabularnewline
\hline 
\textsf{\scriptsize{}$w_{0}$ consists only of letters} & \textsf{\scriptsize{}0.142} & \textsf{\scriptsize{}0.132}\tabularnewline
\hline 
\textsf{\scriptsize{}First token in title} & \textsf{\scriptsize{}0.134} & \textsf{\scriptsize{}0.049}\tabularnewline
\hline 
\textsf{\scriptsize{}$w_{-1}$} & \textsf{\scriptsize{}0.072} & \textsf{\scriptsize{}0.029}\tabularnewline
\hline 
\textsf{\scriptsize{}$w_{0}$ contains hyphen} & \textsf{\scriptsize{}0.071} & \textsf{\scriptsize{}-}\tabularnewline
\hline 
\textsf{\scriptsize{}$w_{0}[0]$ is uppercase} & \textsf{\scriptsize{}0.066} & \textsf{\scriptsize{}0.027}\tabularnewline
\hline 
\textsf{\scriptsize{}Number of characters in $w_{0}$} & \textsf{\scriptsize{}0.064} & \textsf{\scriptsize{}0.096}\tabularnewline
\hline 
\textsf{\scriptsize{}$w_{-1}=$ by} & \textsf{\scriptsize{}0.061} & \textsf{\scriptsize{}-}\tabularnewline
\hline 
\textsf{\scriptsize{}$w_{0}$ lemma} & \textsf{\scriptsize{}0.056} & \textsf{\scriptsize{}0.095}\tabularnewline
\hline 
\textsf{\scriptsize{}$i$ (token position)} & \textsf{\scriptsize{}0.037} & \textsf{\scriptsize{}-}\tabularnewline
\hline 
\textsf{\scriptsize{}$w_{0}[0]$ and $w_{1}[0]$ both uppercase} & \textsf{\scriptsize{}0.036} & \textsf{\scriptsize{}0.078}\tabularnewline
\hline 
\textsf{\scriptsize{}$w_{0}$ is first token in title} & \textsf{\scriptsize{}0.032} & \textsf{\scriptsize{}-}\tabularnewline
\hline 
\textsf{\scriptsize{}$w_{-1}=$ and} & \textsf{\scriptsize{}0.021} & \textsf{\scriptsize{}-}\tabularnewline
\hline 
\textsf{\scriptsize{}$w_{-1}[0]$ uppercase} & \textsf{\scriptsize{}0.014} & -\tabularnewline
\hline 
\textsf{\scriptsize{}$w_{0}$ consists only of digits} & \textsf{\scriptsize{}-} & \textsf{\scriptsize{}0.035}\tabularnewline
\hline 
\textsf{\scriptsize{}$w_{0}$ is uppercase} & \textsf{\scriptsize{}-} & \textsf{\scriptsize{}0.081}\tabularnewline
\hline 
\end{tabular}{\scriptsize \par}

\section{Post Processing \label{SEC:POSTPROCESSING}}

An artifact of employing a sequence labeling algorithm for the task
of extracting brands is that the extracted brand value for any product
title is a substring (possibly empty) of the title. As a consequence,
an incorrect brand name prediction is rarely itself also a brand name.
For example, consider the product title
\begin{quote}
\textsf{\textbf{\scriptsize{}Sea Gull Lighting}}\textsf{\scriptsize{}
Parkfield 3 Light Bath Vanity Light}{\scriptsize \par}
\end{quote}
where the brand name is \emph{`}Sea Gull Lighting\emph{'. }Assume
that we are introducing products from \emph{`}Sea Gull Lighting\emph{'}
for the first time, and thus said brand is not present in the list
of known brands. Further, any substring of this product title consisting
of consecutive tokens is not a brand name. This motivated the following
post processing scheme which significantly boosted precision of the
accepted brand predictions.

\subsection{Normalization\label{SUB:NORMALIZATION}}

As mentioned in section \ref{SEC: CHALLENGES}, a given brand name
may appear in a variety of spellings (for instance missing spaces,
presence/absence of special characters, use of acronyms, etc) in product
titles. In order to be fit to display as a facet, any variations of
a single brand name must be normalized to a unique value. We maintain
a collection of key-value pairs where the key is a brand name variation
and the value is the normalized brand name. The normalized values
are curated by internal analysts. We will call this collection the
\emph{`normalization} \emph{dictionary'.}

For example, consider the following titles corresponding to the brand
`Chenille Kraft'. We highlight the relevant brand tokens in bold.
\begin{itemize}
\item \textsf{\textbf{\scriptsize{}Chenille Kraft }}\textsf{\scriptsize{}Wonderfoam
Magnetic Alphabet Letters, Assorted Colors, 105pk}{\scriptsize \par}
\item \textsf{\textbf{\scriptsize{}THE CHENILLE KRAFT COMPANY}}\textsf{\scriptsize{}
Regular Stems, 6'' X 4Mm, 100/Pack}{\scriptsize \par}
\item \textsf{\textbf{\scriptsize{}Chenillekraft}}\textsf{\scriptsize{}
Round Wood Paint Brush Set - 24 Brush{[}es{]} - Nickel Plated Ferrule
- Wood Handle (CKC5172)}{\scriptsize \par}
\item \textsf{\scriptsize{}Shoplet Best Value Kit - }\textsf{\textbf{\scriptsize{}ChenilleKraft}}\textsf{\scriptsize{}
Assorted Brush Starter Set (CKC5180)}{\scriptsize \par}
\end{itemize}
Thus, for the brand `Chennile Kraft', we add the following entries
to the normalization dictionary:
\begin{enumerate}
\item {\footnotesize{}\{`Chenille Kraft': `Chenille Kraft'\}}{\footnotesize \par}
\item {\footnotesize{}\{`THE CHENILLE KRAFT COMPANY': `Chenille Kraft'\}}{\footnotesize \par}
\item {\footnotesize{}\{`Chenillekraft': `Chenille Kraft'\}}{\footnotesize \par}
\item {\footnotesize{}\{`ChenilleKraft', `Chenille Kraft'\}}{\footnotesize \par}
\end{enumerate}

\subsection{Blacklist}

In addition to the normalization dictionary, we also maintain a list
of terms not known to be brand names. Typically, the list consists
of terms from product titles where no brand name is present. This
list is enlarged whenever the brand extraction process produces an
unacceptable value. We will provide more details in the next section.

\subsection{Manual feedback\label{SUB:MANUAL}}

In this subsection, we consider the case where brand extraction is
done in batch mode. During a single run, every product with a missing
brand value within a category for which a brand facet is applicable
is a candidate for extraction. 

We follow the following procedure to obtain the final result set $R$.
Suppose the brand extraction algorithm runs on $n$ items indexed
$1,2,...,n$. Let the prediction for item $i$ be denoted $\text{brand}(i)$
and denote the normalization dictionary%
\footnote{We use the same symbol to denote both the normalization dictionary
and the normalization function from section \ref{sub:Problem-set-up}.%
} and blacklist as $N$ and $B$ respectively. For each predicted value
not in $N$ we maintain a dictionary $\text{F}$ which tracks the
count of a given value encountered thus far. We will ignore predicted
values that are either not in the normalization dictionary or which
do not meet a predetermined threshold frequency $f$ in a given batch.

\medskip{}

\fbox{\begin{minipage}[t]{0.9\columnwidth}%
for $i=1$ to $n$

\qquad{}if the $\text{brand}(i)\in N$

\qquad{}\qquad{}add $(i,N(\text{brand}(i)))$ to $R$

\qquad{}else

\qquad{}\qquad{}$F(\text{brand}(i))=F(\text{brand}(i))+1$

for brand in $F$

\qquad{}if $F(\text{brand})>f$

\qquad{}\qquad{}Randomly sample $m$ (product title, brand) 

\qquad{}\qquad{}pairs for manual validation of newly 

\qquad{}\qquad{}discovered brands%
\end{minipage}}

\medskip{}
We used the values of $f=30$ and $m=5$.

The sampled (product title, brand) pairs are sent to analysts for
verifying newly discovered brand values. The analyst make the following
decisions for each suggested brand:
\begin{itemize}
\item If the predicted brand value $v_{p}$ is correct for each of the $m$
items:

\begin{enumerate}
\item Suggest a normalized form $v_{n}$ for the brand name.
\item Add the pair $(v_{p},v_{n})$ to $N$ and for any index $i$ with
$\text{brand}(i)=v_{p}$, add $(i,v_{n})$ to $R$.
\end{enumerate}
\item Else

\begin{enumerate}
\item If the predicted value $v_{p}$ is deemed unlikely to be a brand name,
$v_{p}$ is added to $B$ (for instance in the case of unbranded items).
\item If the product title contains a legitimate brand name then the product
title is annotated with this brand name and added to the training
set to build a new model.
\end{enumerate}
\end{itemize}
It should be noted here that a large number of predictions have a
high item count, in some cases of the order of tens of thousands.
Thus, the number of predictions needed to be validated by analysts
is usually a very small fraction of the number of products with predictions.

\section{\label{SEC:RESULTS}Results}

\subsection{Metrics \label{sub:Metrics}}

Now we define the metrics used to measure the performance of the brand
extraction algorithm. Let $n$ be the number of products on which
we run the algorithm. Let $n_{TB}$ be the number of products which
have a true brand value other than \emph{`unbranded'. }Let $n_{PB}$
be the number of products which have a predicted brand value other
than `\emph{unbranded'}. Let $c$ be the number of predictions that
are correct and which have a brand value other than `\emph{unbranded'}.
Then we define 
\[
\text{Precision}=P=\frac{c}{n_{PB}}
\]
and 

\[
\text{Recall}=R=\frac{c}{n_{TB}}.
\]
The motivation for computing precision and recall over only items
with brand value other than `\emph{unbranded' }stems from the fact
that `\emph{unbranded' }value is not displayed in facets and thus
these metrics more accurately represent the quality of the brand facet.
Finally we define $F_{1}$ measure as usual:
\[
F_{1}=\frac{2PR}{P+R}
\]

For diagnostic purposes we also include the metric - `\emph{Label
Accuracy'} which is the fraction of tokens in the data set that are
assigned the correct BIO encoding labels \emph{i.e. }$\{\text{B-brand, I-brand, O}\}$.
Although by itself a high label accuracy does not imply a good sequence
labeling, we can get meaningful information from this metric when
used in combination with precision and recall. 

Additionally, for the LCCRF model, we define Precision and Recall
as functions of the threshold $\theta$ of the conditional probability
of the predicted label sequence \cite{msr}. Let $X_{T}$ denote the
set of all sequences in the test set. For $\tilde{\mathbf{x}}\in X_{T}$,
suppose the LCCRF model predicts a label sequence $\tilde{\mathbf{y}}$
and let the correct label sequence for $\tilde{\mathbf{x}}$ be $\tilde{\mathbf{y}_{t}}$.
Let $\mathbf{o}_{l}$ denote a label sequence of length $l$ in which
every label is `O' (note that this translates to an output brand label
of \emph{`unbranded'). }Let $I[\cdot]$ denote the indicator function.
Then we define
\begin{align*}
 & \text{Precision}(\theta)=P(\theta)=\\
 & \frac{\Sigma_{\mathbf{\tilde{\mathbf{x}}}\in X_{T}}I[(\text{Pr}(\mathbf{\tilde{\mathbf{y}}}|\mathbf{\tilde{\mathbf{x}}})\geq\theta)\text{ and }\mathbf{\tilde{\mathbf{y}}\neq}\mathbf{o}_{|\mathbf{y|}}\text{ and }\tilde{\mathbf{y}}=\tilde{\mathbf{y}_{t}}]}{\Sigma_{\mathbf{\tilde{\mathbf{x}}}\in X_{T}}I[(\text{Pr}(\mathbf{\tilde{\mathbf{y}}}|\tilde{\mathbf{x}})\geq\theta)\text{ and }\mathbf{\tilde{\mathbf{y}}\neq}\mathbf{o}_{|\mathbf{\tilde{\mathbf{y}}|}}]}
\end{align*}

and
\begin{align*}
 & \text{Recall}(\theta)=R(\theta)=\\
 & \frac{\Sigma_{\mathbf{\tilde{\mathbf{x}}}\in X_{T}}I[(\text{Pr}(\mathbf{\tilde{\mathbf{y}}}|\mathbf{\tilde{\mathbf{x}}})\geq\theta)\text{ and }\mathbf{\tilde{\mathbf{y}}\neq}\mathbf{o}_{|\mathbf{y|}}\text{ and }\mathbf{\tilde{\mathbf{y}}}=\tilde{\mathbf{y}_{t}}]}{\Sigma_{\mathbf{\mathbf{\tilde{\mathbf{x}}}}\in X_{T}}I[\tilde{\mathbf{y}_{t}}\mathbf{\neq}\mathbf{o}_{|\mathbf{\tilde{\mathbf{y}_{t}}|}}]}.
\end{align*}
Note that in particular, $P(\theta=0)=P$ and $R(\theta=0)=R$.

\subsection{Model performance \label{sub:Model-results}}

We built a training set consisting of 61,374 product titles from a
selected set of product categories, annotated with a brand attribute
value. We performed a 10-fold cross validation for hyperparameter
selection and computing model metrics.

We compare the results of SP and LCCRF models to a number of baseline
approaches.

\subsubsection{Nearest Neighbors models}

The nearest neighbors model was chosen for comparison for two main
reasons:
\begin{enumerate}
\item The number of distinct brand labels in the training set is 25,851. 
\item The number of times a particular brand label appears in the training
set is between 1-9. The actual distribution of the frequency of the
brand labels in the training set is shown below. 
\end{enumerate}
\begin{figure}[H]
\fbox{\begin{minipage}[t]{0.97\columnwidth}%
\includegraphics[width=0.99\columnwidth]{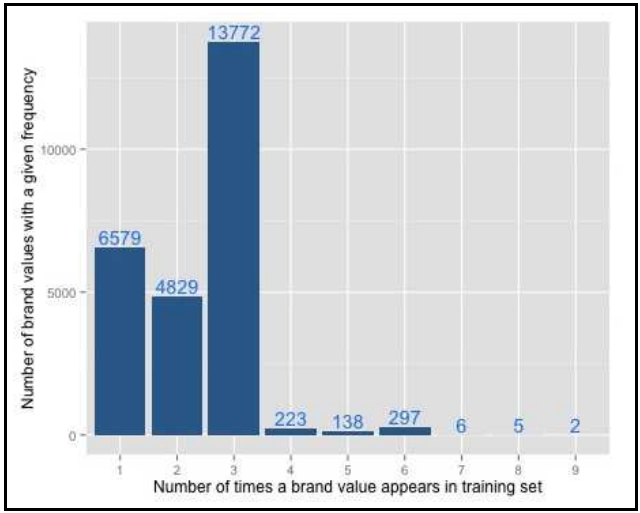}%
\end{minipage}}

\protect\caption{Distribution of label frequencies in the training set}
\end{figure}

Given the large number of classes and since over 95\% of labels appear
3 times or less in the training set, there is very little training
data per class to train an effective model using text classification
techniques.

To obtain the nearest neighbors benchmarks, we used a bag of words
model where each token was encoded using its tf-idf score and the
nearest neighbors were obtained using cosine similarity. Given the
frequency of labels shown above, we compare our results to those of
k nearest neighbors models with k equal to 1 (1NN) and 3 (3NN). For
3NN, in case all three closest neighbors corresponded to distinct
labels, the label with the highest cosine similarity was chosen \emph{i.e.
}the result would coincide with that of 1NN in this case. All results
were obtained using a 10-fold cross validation. 

Note that since some labels appear only once, if such a label occurs
in the test set during cross validation, both 1NN and 3NN would be
incapable of producing the correct result. However, considering one
of the goals of the attribute extraction system is to discover new
attribute values, this situation closely models the reality of the
data.

\subsubsection{Dictionary based approaches}

For dictionary based benchmarks, we compile a lexicon of brands that
appear in the training data with some preprocessing (lowercasing and
tokenization). Given a test title, we perform the same preprocessing
and we search for the presence of any brand from the lexicon in the
title. We show the results from two approaches that were obtained
from the heuristic used to resolve multiple matches:
\begin{enumerate}
\item Dict-max: In case of multiple matches, we choose the match with the
largest number of characters.
\item Dict-first: In case of multiple matches, we choose the match that
appears earliest in the title.
\end{enumerate}

\subsubsection{Hidden Markov Model}

For an input sequence $\mathbf{x}=x_{1},x_{2},...,x_{m}$, and a label
sequence $\mathbf{y}=y_{1},y_{2},...,y_{m}$, a second-order Hidden
Markov Model (HMM) defines the joint probability of the input and
label sequences as follows
\begin{eqnarray*}
 &  & \text{Pr}(x_{1},x_{2},...,x_{m},y_{1},y_{2},...,y_{m})\\
 &  & =\Pi_{i=1}^{m+1}\text{Pr}(y_{i}|y_{i-2},y_{i-1})\Pi_{i=1}^{m}\text{Pr}(x_{i}|y_{i})
\end{eqnarray*}
where $y_{-1},y_{0}=\text{START}$ and $y_{m+1}=\text{STOP}$ are
special labels. The probabilities on the right hand side of the above
equation are parameters of the HMM and can be estimated from a training
set of input and label sequences using maximum likelihood estimation.
Given a test sequence, the label sequence that maximizes the above
joint probability is chosen.

In estimating probabilities for unknown words, we mapped them to one
of several morphological classes similar to \cite{NER}.

\subsubsection{Model comparisons}

The results for all models are shown below. The error margins indicate
two standard errors.

\begin{table}[H]
\begin{centering}
\textsf{\scriptsize{}}%
\begin{tabular}{|c|c|c|c|}
\hline 
 & \textsf{\textbf{\scriptsize{}Precision (\%)}} & \textsf{\textbf{\scriptsize{}Recall (\%)}} & \textsf{\textbf{\scriptsize{}Label Accuracy (\%)}}\tabularnewline
\hline 
\textsf{\textbf{\scriptsize{}LCCRF}} & \textsf{\scriptsize{}91.94$\pm$0.25} & \textsf{\scriptsize{}92.21$\pm$0.25} & \textsf{\scriptsize{}98.44$\pm$0.04}\tabularnewline
\hline 
\textsf{\textbf{\scriptsize{}SP}} & \textsf{\scriptsize{}91.98$\pm$0.21} & \textsf{\scriptsize{}92.18$\pm$0.22} & \textsf{\scriptsize{}98.44$\pm$0.03}\tabularnewline
\hline 
\textsf{\textbf{\scriptsize{}HMM}} & \textsf{\scriptsize{}79.73$\pm$0.28} & \textsf{\scriptsize{}86.13$\pm$0.26} & \textsf{\scriptsize{}92.41$\pm$0.09}\tabularnewline
\hline 
\textsf{\textbf{\scriptsize{}Dict-max}} & \textsf{\scriptsize{}80.91$\pm$0.42} & \textsf{\scriptsize{}77.66$\pm$0.45} & \textsf{\scriptsize{}NA}\tabularnewline
\hline 
\textsf{\textbf{\scriptsize{}Dict-first}} & \textsf{\scriptsize{}79.46$\pm$0.36} & \textsf{\scriptsize{}76.27$\pm$0.41} & \textsf{\scriptsize{}NA}\tabularnewline
\hline 
\textsf{\textbf{\scriptsize{}1NN}} & \textsf{\scriptsize{}69.77$\pm$0.43} & \textsf{\scriptsize{}69.66$\pm$0.46} & \textsf{\scriptsize{}NA}\tabularnewline
\hline 
\textsf{\textbf{\scriptsize{}3NN}} & \textsf{\scriptsize{}65.88$\pm$0.62} & \textsf{\scriptsize{}65.75$\pm$0.64} & \textsf{\scriptsize{}NA}\tabularnewline
\hline 
\end{tabular}
\par\end{centering}{\scriptsize \par}

\medskip{}

\protect\caption{Model metrics for brand extraction}
\end{table}

\begin{figure}[H]
\fbox{\begin{minipage}[t]{0.9\columnwidth}%
\includegraphics[width=0.98\columnwidth]{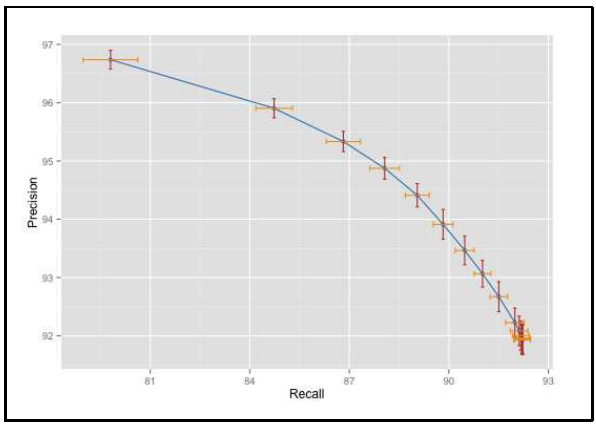}%
\end{minipage}}

\protect\caption{Precision vs Recall for LCCRF model for brand extraction (the error
bars indicate two standard errors)}
\end{figure}

LCCRF and SP outperform the rest of the models and yield similar performance.
The metrics for these models also meet business specifications at
Walmart to power algorithmic solutions to extract product metdata. 

The high label accuracy compared to the precision of the algorithm
indicates a crucial fact that we exploited for post processing. In
a large number of cases even when the predicted sequence labeling
was incorrect, almost all the tokens still had the right label. What
we observed was that the false positive brand value predictions typically
were either subsequences or supersequences of the expected unnormalized
brand. Due to this, we were able to employ the normalization scheme
to improve both precision and recall further. 

We will illustrate this with an example. Consider the following product
title
\begin{quote}
\textsf{\textbf{\scriptsize{}Plum Island Silver }}\textsf{\scriptsize{}SC-007
Sterling Silver Fairy Piece Ear Cuff}{\scriptsize \par}
\end{quote}
that contains the brand name \emph{`}Plum Island Silver\emph{'. }Suppose
the sequence labeling algorithm incorrectly extracts the brand value
as `Plum Island\emph{'. }In such a case, we add the pair (`Plum Island\emph{',
`}Plum Island Silver\emph{')} to the normalization dictionary while
adding the above title in the training set with the correctly annotated
brand. In a subsequent run, it's highly likely that either the algorithm
adapts and correctly predicts the brand or still returns the previous
truncated value. However, even in the latter case, the normalization
scheme will correctly pick the desired normalized value.

\section{Other attributes}

The system is easily extended to a number of other attributes where
we may not have a pre compiled list of possible attribute values or
when such a list is large and may undergo frequent changes. Some examples
include manufacturer specific model numbers and names, fictional characters
that inspire children's toys, sports team and league names for sports
branded apparel and memorabilia, product lines, etc. The approach
presented above is readily amenable for the extraction of such attributes.
Most of the methodology remains similar to that used for brands. We
were able to use the same algorithms, similar set of features, distant
supervision to produce annotated product titles for training and normalization
schemes for standardization of variations of attributes values to
build models with high precision and recall for such attributes.

\section{Real world performance}

We used the approach presented above to extract brands for a subset
of products in selected departments. Early results from algorithmic
extraction were manually validated by analysts. The normalization
dictionary was regularly updated based on analyst feedback. For the
manually validated batches, post normalization precision varied between
92\%-95\% showing an improvement between 1\%-3\% over algorithmically
extracted values. Sample validations of the products categorized as
unbranded revealed that over 98\% of those were indeed true negatives.
The algorithm was able to `discover' over a thousand brand values
which were not part of the Walmart brand database at that point.

We estimated the impact of brand value extraction on product discoverability
using the R package \emph{CausalImpact }\cite{causalimpact}. For
the following analysis, the brand attribute value was extracted and
used to augment product data for 272,697 products on the same day.
We measured the impressions for this set of products during 44 days
prior and 27 days following \footnote{these choices were motivated by the choice of series to create a synthetic control}
this intervention conditioned on the brand facet being applied. The
results of the analysis are shown below.

\begin{figure}[H]
\fbox{\begin{minipage}[t]{0.95\columnwidth}%
\includegraphics[width=0.99\columnwidth]{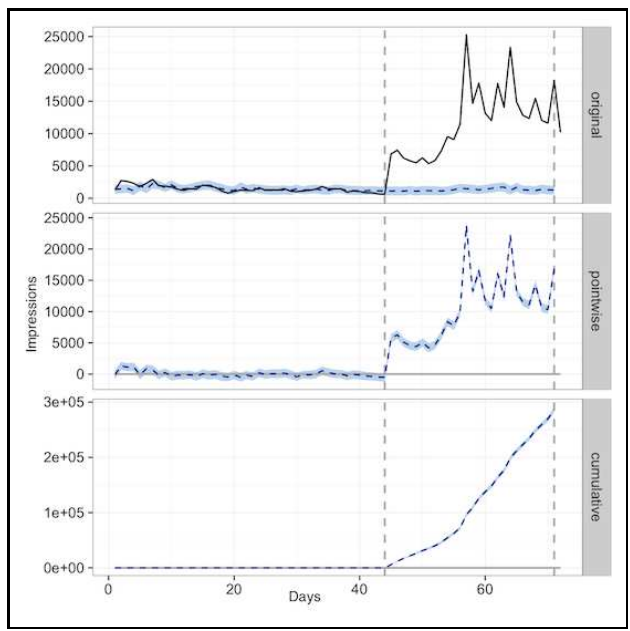}%
\end{minipage}}

\protect\caption{Daily impressions for products for which brand attribute value was
extraced}
\end{figure}

The plot consists of three panels. The solid line in the top panel
shows the actual daily impressions for the set of products. The first
vertical dotted line indicates the day of intervention (products augmented
with brand metadata). The second vertical line marks the end of the
measurement. The blue dotted time series in the first panel is the
predicted counterfactual - how the impressions would have tracked
in the absence of the intervention. The second panel shows the difference
between the observed impressions and the counterfactual predictions
while the final plot shows the cumulative difference. The synthetical
control was constructed from products from similar categories for
which no brand metadata was augmented. The model predicts over 250K
additional impressions during the test period for the set of items
for which the brand metadata was supplied. Similar analysis for other
attributes for which attribute extraction was employed typically suggests
a positive impact on impressions and depending on the attribute also
on other metrics of interest such as clicks, add to cart rate and
orders.

So far, the attribute extraction system has been deployed for extracting
over 20 attributes. We build a separate model for each attribute for
several reasons. A given attribute is typically applicable to only
a particular set of categories (for e.g. brand may not be an important
attribute for books while sports team may not be applicable for electronic
tablet devices). Building a separate model for each attribute allows
us to independently train each model and tune the performance based
on business needs. The size of the training set varies between a few
hundred titles to tens of thousands depending on the attribute.

\section{Conclusion}

In this paper, we introduced the problem of extracting attribute values
from product titles in order to augment product metdata for eCommerce
catalogs. We discussed the importance of attribute extraction for
retail websites. We examined the challenges associated with this task
especially when the product catalog is large. Attribute extraction
from product titles was modeled as a sequence labeling problem. We
also illustrated a method to leverage existing products tagged with
attribute values to build the initial training data set without manual
labeling. The experimental results show that SP or LCCRF models combined
with a curated normalization scheme provide an efficacious mechanism
for tagging products with certain attribute values with high precision
and recall. 

\bibliographystyle{plain}
\bibliography{references}

\end{document}